\documentclass[lettersize,journal, twoside]{IEEEtran}
\usepackage{amsmath,amsfonts}
\usepackage{algorithmic}
\usepackage{algorithm}
\usepackage{array}
\usepackage[caption=false,font=normalsize,labelfont=sf,textfont=sf]{subfig}
\usepackage{textcomp}
\usepackage{stfloats}
\usepackage{url}
\usepackage{verbatim}
\usepackage{graphicx}
\usepackage{multirow} 
\usepackage{multicol}
\usepackage[hidelinks]{hyperref}
\usepackage{cite}
\usepackage{tikz,xcolor}
\usepackage{bm}
\usepackage{fancyhdr}
\usepackage{booktabs}
\usepackage{lettrine}
\usepackage{verbatim}
\usepackage{threeparttable}
\usepackage{makecell}
\usepackage{xcolor}
\usepackage[T1]{fontenc}
\usepackage{cite}
\usepackage{balance}

\usepackage{mathbbol}
\usepackage{amssymb} 
\usepackage[T1]{fontenc}
\DeclareUnicodeCharacter{0301}{\'{e}}
\usepackage{braket}
\usepackage{stmaryrd}

\def\x{{\mathbf x}}

\definecolor{lime}{HTML}{A6CE39}
\DeclareRobustCommand{\orcidicon}{
	\begin{tikzpicture}
		\draw[lime, fill=lime] (0,0)
		circle[radius=0.16]
		node[white]{{\fontfamily{qag}\selectfont \tiny \.{I}D}}; \draw[white, fill=white] (-0.0625,0.095) circle [radius=0.007];
	\end{tikzpicture}
	\hspace{-2mm}
}
\foreach \x in {A, ..., Z}{%
	\expandafter\xdef\csname orcid\x\endcsname{\noexpand\href{https://orcid.org/\csname orcidauthor\x\endcsname}{\noexpand\orcidicon}}
}

\begin{document}

\title{Iterative Feedback Network for Unsupervised \protect\\Point Cloud Registration}

\author{Yifan Xie\orcidA{}, Boyu Wang\orcidD{}, Shiqi Li\orcidB{} and Jihua Zhu\orcidC{}
\thanks{This work was supported in part by the Key Research and Development Program of Shaanxi Province under Grant 2021GY-025 and Grant 2021GXLH-Z-097. (Corresponding author: Jihua Zhu.)}
\thanks{The authors are with the School of Software Engineering, Xi'an Jiaotong University, Xi'an 710000, China(e-mail:xieyifan@stu.xjtu.edu.cn; 1415194648@stu.xjtu.edu.cn; lishiqi@stu.xjtu.edu.cn; zhujh@xjtu.edu.cn). Code will be available at https://github.com/IvanXie416/IFNet.}
\thanks{Digital Object Identifier (DOI): see top of this page.}
}

\markboth{IEEE ROBOTICS AND AUTOMATION LETTERS. PREPRINT VERSION. JANUARY 2024}%
{ Xie \MakeLowercase{\textit{et al.}}: Iterative Feedback Network for Unsupervised Point Cloud Registration}


\maketitle

\begin{abstract}
As a fundamental problem in computer vision, point cloud registration aims to seek the optimal transformation for aligning a pair of point clouds.
In most existing methods, the information flows are usually forward transferring, thus lacking the guidance from high-level information to low-level information. 
Besides, excessive high-level information may be overly redundant, and directly using it may conflict with the original low-level information. 
In this paper, we propose a novel Iterative Feedback Network (IFNet) for unsupervised point cloud registration, in which the representation of low-level features is efficiently enriched by rerouting subsequent high-level features.
Specifically, our IFNet is built upon a series of Feedback Registration Block (FRB) modules, with each module responsible for generating the feedforward rigid transformation and feedback high-level features. These FRB modules are cascaded and recurrently unfolded over time. 
Further, the Feedback Transformer is designed to efficiently select relevant information from feedback high-level features, which is utilized to refine the low-level features.
What’s more, we incorporate a geometry-awareness descriptor to empower the network for making full use of most geometric information, which leads to more precise registration results. 
Extensive experiments on various benchmark datasets demonstrate the superior registration performance of our IFNet.
\end{abstract}

\begin{IEEEkeywords}
	3D point clouds,
	point cloud registration, 
        feedback mechanism,
	attention mechanism.
\end{IEEEkeywords}

\section{Introduction}
\IEEEPARstart{T}{th} rapid development of modern information technology and graphics has resulted in the widespread application of 3D reconstruction technology in various fields, including augmented reality~\cite{billinghurst2015survey}, simultaneous localization and mapping (SLAM)~\cite{ding2019deepmapping}, and autonomous driving~\cite{geiger2012we}. 
One of the most crucial and challenging aspects of the 3D reconstruction process is 3D point cloud registration~\cite{huang2021comprehensive}. This step involves predicting a rigid 3D transformation and aligning the source point cloud with the target point cloud. 
Due to occlusion and noise, point cloud registration continues to be a challenging problem in real-world applications.

\begin{figure}
	\centering
	\includegraphics[width=0.9\columnwidth]{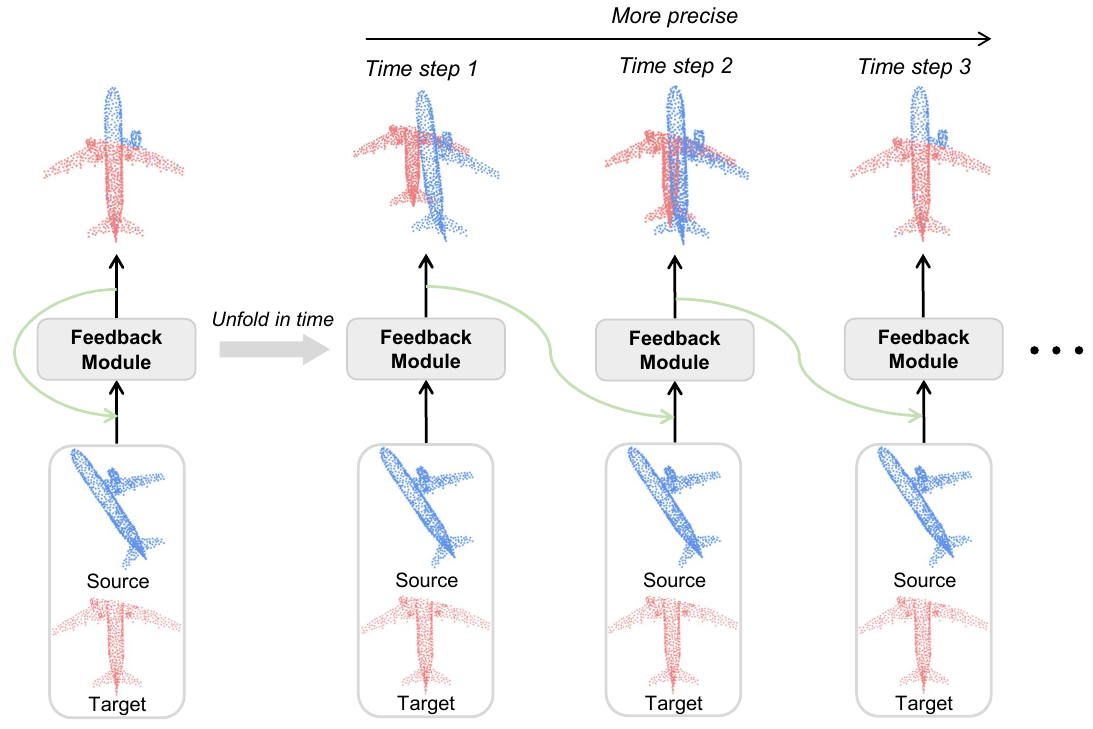} 
	\setlength{\belowcaptionskip}{-0.5cm}
	\caption{The illustrations of the feedback mechanism in our IFNet. Green lines denote the feedback information.
	}
	\label{overview}
	\vspace*{-0.6\baselineskip}
\end{figure}

Traditional methods, such as Iterative Closest Point (ICP)~\cite{besl1992method} and its variants~\cite{yang2015go,rusinkiewicz2019symmetric}, are commonly employed for point cloud registration. However, these methods have limitations: they are sensitive to the initial position of registration and struggle to handle point cloud registration tasks with noise or low overlap.
With the advancement of deep learning in 3D vision tasks, an increasing number of researchers have been exploring the application of learning-based methods to point cloud registration tasks~\cite{aoki2019pointnetlk,wang2019deep,yew2022regtr}, yielding promising and excellent results.
However, collecting the ground truth transformations is both expensive and time-consuming, which can significantly escalate the training cost and impede their practical application in real-world scenarios.
To overcome this limitation, FMR~\cite{huang2020feature} uses point cloud reconstruction for feature extraction with poor registration precision on partial or noisy data. 
All of these methods are iterative and feedforward, and as the iterations proceed, the results of subsequent iterations will be essentially better than the results of earlier iterations. It is natural to raise a question: 
\textit{Could high-level information in subsequent iterations guide the learning of low-level information in earlier iterations?}

In cognition theory, feedback connections linking cortical visual areas can transmit response signals from higher-order areas to lower-order areas~\cite{gilbert2007brain}. Drawing inspiration from this phenomenon, recent studies~\cite{tomar2022fanet,yan2022fbnet} have incorporated the feedback mechanism into network architectures. In these architectures, the feedback mechanism operates in a top-down manner, propagating high-level information back to previous layers and refining low-level information. 
{Specifically, high-level information refers to abstract and generalized information obtained through feature extraction, while low-level information refers to specific and underlying information that contains more details.}
As shown in Fig.~\ref{overview}, we apply the feedback mechanism to the 3D point cloud registration task. 
By cascading multiple feedback modules and unfolding them recurrently across time, the module leverages high-level information from previous time steps to enrich the present low-level information. 

In this paper, we introduce the iterative feedback network (IFNet) for unsupervised point cloud registration, which is the first feedback-based network designed specifically for this task.
The IFNet is composed of multiple Feedback Registration Block (FRB) modules. In each FRB module, high-level features are utilized to enhance the learning process of the low-level features. By leveraging the valuable contextual information from the high-level features, the low-level features become more representative, ultimately improving the network's overall representation capability. 
The question of how to effectively utilize high-level features to guide the low-level features is an important aspect that warrants further investigation. To address this, we propose the Feedback Transformer, which adaptively selects and enhances relevant high-level information to refine the low-level features.
Additionally, we incorporate a geometric-aware descriptor to make our network more sensitive to geometric information. By the stacking of FRB modules, the outputs are gradually refined across time steps, ultimately leading to the precise estimation of rigid transformations.

To summarize, the contributions of our paper include: 
\begin{itemize}
	
\item We propose a novel iterative feedback network (IFNet) for unsupervised point cloud registration, which progressively refines the registration results at each time step and shows superior performance on various benchmark datasets. 
\item We introduce the Feedback Transformer to facilitate the integration of high-level information into the learning process of low-level features.  
\item A geometry-aware descriptor is proposed as a positional embedding, enabling the network to fully utilize geometric information.

\end{itemize}

\section{RELATED WORK}
\subsection{Traditional Point Cloud Registration}
Iterative Closest Point (ICP)~\cite{besl1992method} is a widely used traditional point cloud registration algorithm that aims to align two or more point clouds by iteratively minimizing the distance between corresponding points. 
Despite its simplicity, the ICP algorithm has some limitations. It is sensitive to initial alignment and can get trapped in local minima. It also assumes that the correspondences are accurate and that the point clouds have sufficient overlap. Researchers have proposed extensions and variations of ICP to address these limitations, such as Go-ICP~\cite{yang2015go}, Symmetric ICP~\cite{rusinkiewicz2019symmetric} and so on.
Additionally, RANSAC-based methods~\cite{le2019sdrsac} have also demonstrated effective registration results.

\subsection{Learning-Based Point Cloud Registration}
With the remarkable achievements of deep learning in image processing, researchers have turned their focus towards learning-based point cloud registration methods. 
PointNetLK~\cite{aoki2019pointnetlk} integrates a modified Lucas Kanade algorithm~\cite{baker2004lucas} into PointNet~\cite{qi2017pointnet}, enabling iterative alignment of input point clouds. 
DCP~\cite{wang2019deep} combines Dynamic Graph CNN~\cite{wang2019dynamic} and attention mechanism~\cite{vaswani2017attention} to extract features, and employs pointer networks to predict soft matches between point clouds. 
IDAM~\cite{li2020iterative} introduces a two-stage point elimination technique to aid in generating partial correspondences. 
{PREDATOR~\cite{huang2021predator} projects the features as an overlap score, which can be interpreted as the probability that a point lies in the overlap region.}
IMFNet~\cite{huang2022imfnet} uses cross-modal features for point cloud registration on real datasets.
FINet~\cite{xu2022finet} utilizes a two-branch structure, allowing for separate handling of rotations and translations.
{UDPReg~\cite{mei2023unsupervised} predicts the distribution-level correspondences while considering the mixing weights of Gaussian mixture models to effectively handle partial point cloud registration.}

In recent years, there has been significant research focused on unsupervised point cloud registration due to the absence of ground truth in real-world scenarios. 
CEMNet ~\cite{jiang2021sampling} introduces a differentiable CEM (Cross-Entropy Method) module to enhance the discovery of optimal solutions. 
RIENet ~\cite{shen2022reliable} employs a learnable graph representation to capture geometric disparities between source and pseudo-target neighborhoods. 
GSRNet~\cite{wang2023robust} utilizes an attention module based on the geometric spatial feature differences for unsupervised registration.
While all of these methods achieve remarkable registration accuracy, they are forward-transferring in nature, neglecting the potential influence of high-level information on the learning of low-level information during the registration process.

\subsection{Feedback Mechanism}
The incorporation of a feedback mechanism in deep networks enables low-level features to become more representative and informative by propagating high-level information from deep layers to shallow layers. While this approach has been extensively utilized in various 2D image visualization domains~\cite{sam2018top,conf/bmvc/LiLLJLY19,tomar2022fanet}, its applications in 3D have been limited. 
To address this gap and explore the application of the feedback mechanism on 3D point clouds, we propose a novel attention mechanism-based module. This module refines the low-level point cloud features by incorporating information from high-level point cloud features, making it distinct from previous methods.

\begin{figure*}
	\centering
	\includegraphics[width=0.9\textwidth]{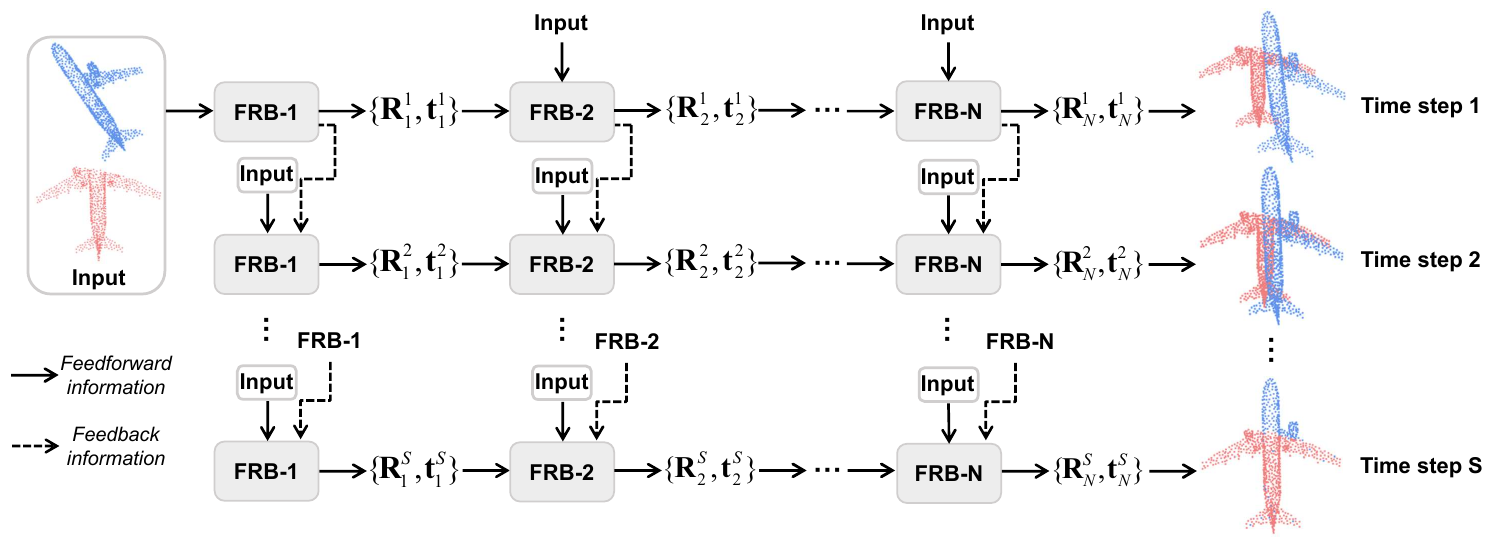} 
	\caption{The overall architecture of IFNet consists of multiple Feedback Registration Block (FRB) modules. Each FRB module generates the feedback information and rigid transformation $\{\mathbf{R}_n^s,\mathbf{t}_n^s\}$, where $n$ represents the number of iterations in the spatial domain and $s$ represents the time step. Additionally, the weight parameters of the FRB modules are shared across time steps.}
	\label{network}
	\vspace{-.3cm}
\end{figure*}

\section{METHOD}

In this section, we demonstrate our unsupervised point cloud registration network, IFNet, which is based on an iterative feedback mechanism. 
The overall architecture consists of multiple Feedback Registration Block (FRB) modules, is presented in Fig.~\ref{network}.
{The internal construction of the FRB module is inspired by ~\cite{shen2022reliable}, on which we design the Feedback Transformer to enhance the integration of high-level features into the learning process of low-level features.}
The stacked FRB modules determine the rigid transformation $\{\mathbf{R}, \mathbf{t}\}$ based on both the initial input and the outputs from the previous FRB, where $\mathbf{R} \in {SO}(3)$ is a rotation matrix and $\mathbf{t} \in \mathbb{R}^3$ is a translation vector. The feedback connections on these FRB modules reroute high-level hybrid information to enhance low-level point features. With the help of the iterative feedback mechanism, the FRB modules can gradually refine the registration results step by step.

\subsection{Iterative Feedback Mechanism}

Most previous point cloud registration methods~\cite{besl1992method, rusinkiewicz2019symmetric, xu2022finet} adopt an iterative approach to progressively refine the registration results for higher accuracy. 
Moreover, a feedback mechanism is often used in the field of 2D images~\cite{zamir2017feedback, conf/bmvc/LiLLJLY19, hu2023high} that enhances low-level features by propagating high-level information to shallower layers.
By combining these two approaches, we propose IFNet, a point cloud registration network that leverages iterative refinement and feedback mechanisms. This integration allows us to achieve more representative and informative low-level features through the utilization of high-level information.

As depicted in Fig.~\ref{network}(from left to right), IFNet comprises $N$ stacked Feedback Registration Block (FRB) modules and establishes multiple feedback connections to capture more effective information. The information initially flows from the initial input to the stacked FRB modules in a feedforward manner. Each FRB module takes the output of the previous FRB and the initial input as inputs, then refines the rigid registration results to be more precise. 

In addition, the information also undergoes a feedback process, flowing from the high layer to the low layer within the same FRB module. We cascade multiple FRB modules and recurrently unfold them across time, as illustrated in Fig.~\ref{network}(from top to down). 
In the $n$-th FRB module at time step $s$, the high-layer features from the previous time step $s - 1$ are rerouted and utilized for the present step's feature learning through the feedback connection. 
We can reasonably assume that high-level features at time step $s - 1$ contain fine-grained information that can refine the low-level features to be more representative and informative at the present time step $s$.
Consequently, the stacked FRB modules can progressively refine registration results as they unfold across time steps.

\begin{figure*}
	\centering
	\includegraphics[width=0.9\textwidth]{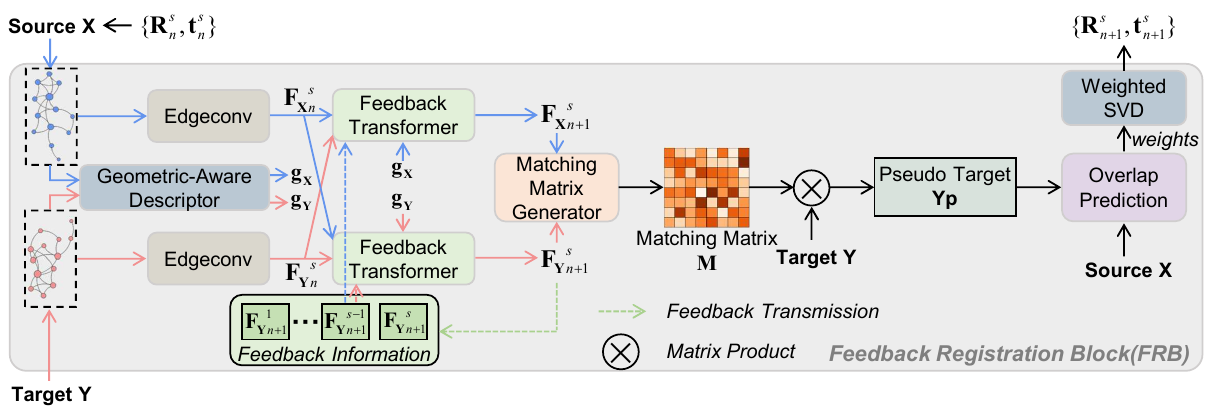} 
	\caption{The detailed structure of the Feedback Registration Block (FRB).}
	\label{FRB}
\end{figure*}

\begin{figure}
	\centering
	\includegraphics[width=0.4\textwidth]{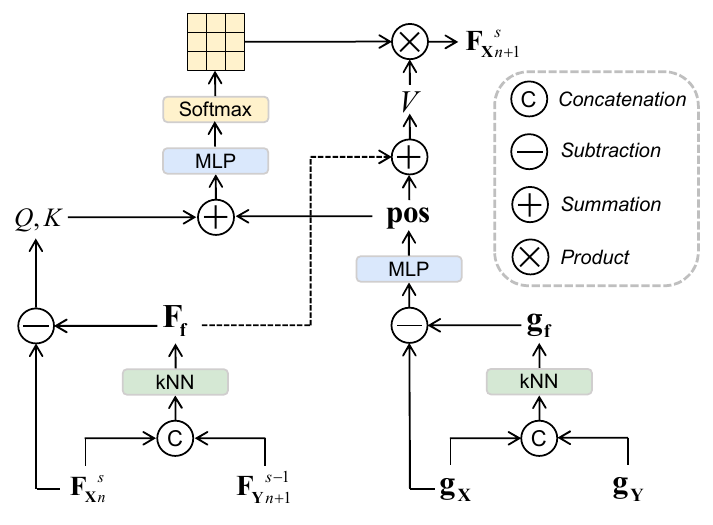} 
	\caption{The pipeline of the Feedback Transformer.
	}
	\label{FT}
	\vspace{-.2cm}
\end{figure}

\subsection{Feedback Registration Block}
The purpose of the Feedback Registration Block (FRB) module is to generate a precise rigid transformation. Fig.~\ref{FRB} illustrates the various components of the FRB, which include Feature Extraction, Geometry-Aware Descriptor, Feedback Transformer, Matching Matrix Generator and Overlap Prediction. 
We will detail in turn below.

\emph{Feature Extraction and Geometry-Aware Descriptor.} 
For feature extraction, we consider each point in the point clouds $\mathbf{X}$ and $\mathbf{Y}$ as a vertex in a graph. We then calculate the pointwise feature using the EdgeConv~\cite{wang2019dynamic} operation. 

To make the model more sensitive to geometric features, we propose the geometric-aware descriptor. Specifically, we first employ the k-nearest neighbor algorithm to locate the two nearest points $\mathbf{p}_{i1}$, $\mathbf{p}_{i2}$ of vertex $\mathbf{p}_{i}$.
Subsequently, we can construct the geometry-aware descriptor $\mathbf{g}_{i}$ via edges, edge lengths and normal:
\begin{equation}
\footnotesize
\label{eq1}
\mathbf{g}_{i} = \mathrm{cat}\left[\mathbf{p}_i,\mathbf{edge}_1,\mathbf{edge}_2,\mathbf{length}_1,\mathbf{length}_2,\mathbf{normal}\right],
\end{equation}
where $\mathrm{cat} [\cdot]$ denotes the concatenation,
{$\mathbf{edge}_j=\mathbf{p}_{ij}-\mathbf{p}_i$, $\mathbf{length}_j=\mid \mathbf{edge}_j\mid$ and $\mathbf{normal}=\mathbf{edge}_1\times \mathbf{edge}_2$.
}


To expand the geometric features of each point, we incorporate length, edge, and normal information. Once we obtain the geometry-aware descriptor for each point, we utilize it to compute the positional embedding in the subsequent Feedback Transformer. 
This enables the model to leverage the enriched geometric information.

\emph{Feedback Transformer.} 
To facilitate the integration of high-level information into the learning process of low-level features, we develop the Feedback Transformer. As depicted in Figure~\ref{FT}, we combine the low-level features $\mathbf{F_X}^{s}_{n}$ from the source point cloud $\mathbf{X}$ at time step $s$ with the high-level features $\mathbf{F_Y}^{s-1}_{n+1}$ from the target point cloud $\mathbf{Y}$ at time step $s - 1$, after which the fused features $\mathbf{F_f}$ are obtained using the k-nearest neighbor algorithm:
\begin{equation}
\label{eq3}
\mathbf{F_f} =  \mathrm{kNN}(\mathrm{cat}[\mathbf{F_X}^{s}_{n},\mathbf{F_Y}^{s-1}_{n+1}]).
\end{equation}
Additionally, we can obtain the positional embedding $\mathbf{pos}$:
\begin{equation}
\label{eq4}
\mathbf{pos} = \mathbf{g_X} - ( \mathrm{kNN}(\mathrm{cat}[\mathbf{g_X},\mathbf{g_Y}])).
\end{equation}
We can then utilize the fused features $\mathbf{F_f}$ to guide the model in learning a better feature representation of $\mathbf{F_X}^{t}_{n}$. The entire process can be outlined as follows:
\begin{equation}
\footnotesize
\label{eq5}
\mathbf{F_X}^{s}_{n+1} = \mathrm{softmax} (\mathrm{MLP} ({\mathbf{F_X}^{s}_{n}} - \mathbf{F_f}) + \mathbf{pos}) \cdot (\mathbf{F_f} + \mathbf{pos}).
\end{equation}
The neighborhood information is constructed using the k-nearest neighbor algorithm, allowing $\mathbf{F_X}^{s}_{n}$ to query salient information from $\mathbf{F_f}$ and enrich its features.

When the time step is $0$, there is no previous time, so instead of using the feedback mechanism, $\mathbf{F_X}^{0}_{n}$ and $\mathbf{F_Y}^{0}_{n}$ are used directly as inputs. 
With the Feedback Transformer, it is reasonable to speculate that the features of the keypoints, such as overlap points and edge points, will become more prominent, which is helpful for the registration task. We employ the high-level target point cloud features to guide the low-level source point cloud features, making the source point cloud features more focused on learning the regions of keypoints. It is worth noting that the source point cloud undergoes a rigid transformation at the input of each FRB module, so we do not utilize the high-level source point cloud features to guide the low-level target point cloud features.

\begin{figure}
	\centering
	\includegraphics[width=0.45\textwidth]{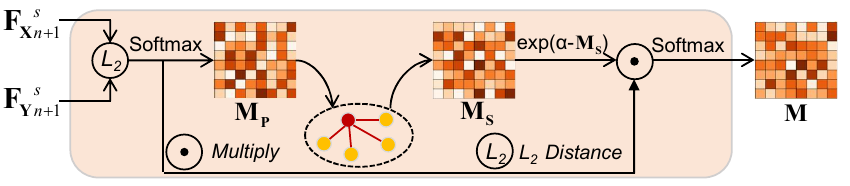} 
	\caption{The detailed structure of the Matching Matrix Generator.
	}
	\label{sup1}
	\vspace{-.2cm}
\end{figure}

\emph{Matching Matrix Generator.} 
The matching matrix generator is used to generate matching matrix with high quality correspondences. The detailed structure is shown in Fig.~\ref{sup1}.
After obtaining the features $\mathbf{F_X}^{s}_{n+1}$ and $\mathbf{F_Y}^{s}_{n+1}$, the feature differences are used to obtain the preliminary matching matrix $\mathbf{M_P}$.
Based on the preliminary matching matrix, we can calculate the neighboring score of each point:
\begin{equation}
\mathbf{M_S}_{i,j}=\frac1K\sum_{\mathbf{x}_{i^{\prime}}\in\mathcal{N}_{\mathbf{x}_i}\mathbf{y}_{j^{\prime}}\in\mathcal{N}_{\mathbf{y}_j}}\mathbf{M_P}_{i^{\prime},j^{\prime}},
\end{equation}
where $\mathcal{N}_{\mathbf{x}_i}$ denotes the neighboring point set of $\mathbf{x_i}$, and $K$ is the number of neighboring points. 
The neighboring score matrix $\mathbf{M_S}$ defined here is negatively correlated with the corresponding correlation degree of the point pairs. Therefore, we use the negative exponential function to expand the differentitation degree of the correspondence relationships:
\begin{equation}
\mathbf{M_S'}_{i,j}=\exp{(\alpha-\mathbf{M_S}_{i,j})}*\mathbf{D}_{i,j},
\end{equation}
where $\mathbf{D}_{i,j}$ denotes the Euclidean distance between the features of point $\mathbf{x_i}$ and $\mathbf{y_j}$, $\alpha$ is used to regulate the effects of the neighboring score. Finally, we can compute the final matching matrix using the softmax operation:
\begin{equation}
\mathbf{M}_{i,j}=\mathrm{softmax}\left(\begin{bmatrix}-\mathbf{M_S'}_{i,1},\ldots,-\mathbf{M_S'}_{i,N}\end{bmatrix}\right)_j,
\end{equation}
where $N$ is the total number of points in the point cloud.

\begin{figure}
	\centering
	\includegraphics[width=0.45\textwidth]{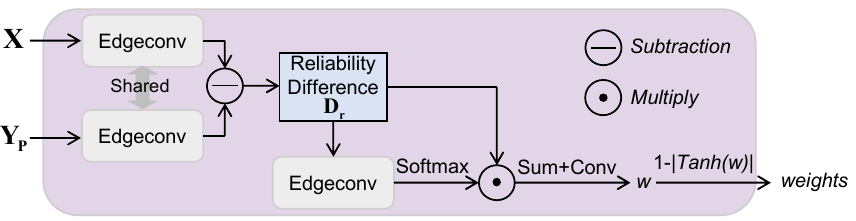} 
	\caption{The specific structure of the Overlap Prediction.
	}
	\label{sup2}
	\vspace{-.2cm}
\end{figure}

\emph{Overlap Prediction.} 
In this section, we introduce the overlap prediction module to compute the weights and thus infer the overlap regions between two point clouds.
The specific structure of the overlap prediction is shown in Fig.~\ref{sup2}.
Specifically, we first project the target point cloud using the final matching matrix $\mathbf{M}$ to obtain the pseudo target point cloud $\mathbf{Y_P}$. The pseudo target point cloud should have similar features to the corresponding source neighborhood.
Then, EdgeConv~\cite{wang2019dynamic} $v$ is utilized to construct the graph representation of neighboring points, denoted as $\mathbf{e}^{x}_{i,k}$ and $\mathbf{e}^{y_p}_{i,k}$. And we use the subtraction operation to obtain the reliability difference $\mathbf{D_r}$:
\begin{equation}
\mathbf{D_r}_{i,k} = v(\mathbf{e}^{x}_{i,k}) - v(\mathbf{e}^{y_p}_{i,k}).
\end{equation}

Subsequently, we learn the attention coefficients of the reliability difference through another EdgeConv $u$:
\begin{equation}
\tau_{i,k}=\text{softmax}\left(\text{cat}\left[u\left(\mathbf{D_r}_{i,k}\right)\right]_{1}^{K}\right),
\end{equation}
where $K$ denotes the number of neighboring points.
Finally, we sum the reliability differences weighted by the attention coefficients and learn to obtain the pseudo-corresponding overlap weights:
\begin{equation}
{weights}_i=1-\tanh\left(\left|f\left(\sum_{k=1}^K\tau_{i,k}*\mathbf{D_r}_{i,k}\right)\right|\right),
\end{equation}
where $f (\cdot)$ denotes the one-dimensional convolution operation. 
If the weight is larger, it means that the points tend to be overlapping points.

\begin{table*}[t]
\caption{Results of different methods on ModelNet40, where $\left(\star\right)$, $\left(\circ\right)$, and $\left(\triangleleft\right)$ denote traditional, supervised and unsupervised methods, respectively. The boldface and underline indicate the best and second-best performance, respectively.}
\setlength{\abovecaptionskip}{0pt}
\setlength{\belowcaptionskip}{0pt}
\begin{center}
        {    \resizebox{1\linewidth}{!}{
                \begin{tabular}{lcccc|cccc|cccccccc}
                    \toprule
                    \multicolumn{1}{c}{\multirow{2}{*}{Method}} &\multicolumn{4}{c|}{(a) Same} & \multicolumn{4}{c|}{(b) Unseen} & \multicolumn{4}{c}{(c) Noise} \\
                    \cmidrule{2-3}   \cmidrule{4-5} \cmidrule{6-7}   \cmidrule{8-9} \cmidrule{10-11}   \cmidrule{12-13} \cmidrule{14-15}
                    &RMSE(\textbf{{R}})&MAE(\textbf{{R}})&RMSE(\textbf{{t}})&MAE(\textbf{{t}})&RMSE(\textbf{{R}})&MAE(\textbf{{R}})&RMSE(\textbf{{t}})&MAE(\textbf{{t}})
                    &RMSE(\textbf{{R}})&MAE(\textbf{{R}})&RMSE(\textbf{{t}})&MAE(\textbf{{t}})\\
                    \midrule
                    ICP $\left(\star\right)$& 33.6842& 25.0537& 0.2912& 0.2524 &  34.2744&  25.6378&  0.2924& 0.2519& 30.5018&  24.0121& 0.2391& 0.2184  \\
                    FGR $\left(\star\right)$&3.7055 & 0.5972&0.0088  &0.0020& 3.1251&  0.4469&  0.0074& 0.0013& 4.5798&  1.2173&  0.0186& 0.0051  \\
                    SDRSAC $\left(\star\right)$&3.9173 & 2.7956 & 0.0121 & 0.0102 &4.2475&  3.0144&  0.0139&  0.0121& 3.8351&  3.0619& 0.0142&  0.0128 \\
                    DCP $\left(\circ\right)$&  6.6498&  4.8472& 0.0273& 0.0215&  9.8374& 6.6458&  0.0338&  0.0252&  16.4068& 13.3563&  0.1120& 0.0887\\
                    IDAM $\left(\circ\right)$&2.4612&0.5618 &0.0167 &0.0035&3.0425&0.6160 & 0.0197&0.0048&13.2725&11.1256 & 0.0831& 0.0661  \\
                    FINet $\left(\circ\right)$&  1.4631& 0.6427& 0.0112&  0.0068 & 2.3915& 0.8015& 0.0105& 0.0045& 2.5171& 1.7000& 0.0163& 0.0124\\
                    FMR $\left(\triangleleft\right)$& 9.0997&3.6497& 0.0204 &0.0101&9.1322 &3.8593 & 0.0233  &0.0113&8.3698 &3.9390& 0.0291 &0.0149\\
                    CEMNet $\left(\triangleleft\right)$& 1.5018&  0.1385& 0.0009& \underline{0.0001}& 1.1013& 0.0804&0.0020& 0.0002& 3.2477& 0.4047& 0.0076& 0.0013 \\
                    RIENet $\left(\triangleleft\right)$ &\underline{0.0246} &\underline{0.0120} & \underline{0.0001} & \textbf{0.0000} &\underline{0.0298} &\underline{0.0110} & \underline{0.0002} & \underline{0.0001} &\underline{0.1003} & \underline{0.0386}& \underline{0.0004} & \underline{0.0002}\\
                    \midrule
                    Ours $\left(\triangleleft\right)$& \textbf{0.0016}& \textbf{0.0007}& \textbf{0.0000}&\textbf{0.0000}& \textbf{0.0013}& \textbf{0.0006}& \textbf{0.0000}& \textbf{0.0000} & \textbf{0.0220}&\textbf{0.0071}& \textbf{0.0000}& \textbf{0.0000}\\
                    \bottomrule
            \end{tabular}}
        }
\end{center}
\label{table1}
\vskip -5pt
\end{table*}

\subsection{Loss Functions}
\emph{Global Registration Loss.}
We train our model using a loss function based on a distance measure. The global registration loss is defined as:
\begin{equation}
    \mathcal{L}_{gr}=\sum_{\mathbf{x}^{\prime}\in \mathbf{X}^{\prime}} \gamma \left(\min_{\mathbf{y}\in \mathbf{Y}}\|\mathbf{x}^{\prime}-\mathbf{y}\|_2^2\right)+\sum_{\mathbf{y}\in \mathbf{Y}} \gamma \left(\min_{\mathbf{x}^{\prime}\in \mathbf{X}^{\prime}}\|\mathbf{y}-\mathbf{x}^{\prime}\|_2^2\right),
\end{equation}
where $\mathbf{X}^{\prime}$ represents the transformed source point cloud using the predicted transformation and $\gamma$ is the huber function.

\emph{Neighborhood Consistency Loss.}
We can obtain the ${k}$ point pairs with the highest weights based on the overlap prediction. Then, we can obtain the overlapping point clouds $\mathbf{X}_o\in \mathbb{R}^{{k}\times3}$ and $\mathbf{Y}_o\in \mathbb{R}^{{k}\times3}$. So we can construct the neighborhood consistency loss using the transformation $\left\{\mathbf{R}, \mathbf{t}\right\}$:
\begin{equation}
\mathcal{L}_{nc}=\sum_{\mathbf{x}_i\in\mathbf{X}_o,\mathbf{y}_i\in\mathbf{Y}_o}\sum_{\mathbf{p}_j\in\mathcal{N}_{\mathbf{x}_i},\mathbf{q}_j\in\mathcal{N}_{\mathbf{y}_i}}\|{\mathbf{R}}\mathbf{p}_j+{\mathbf{t}}-\mathbf{q}_j\|_2,
\end{equation}
where $\mathcal{N}_{\mathbf{x}_i}$ and $\mathcal{N}_{\mathbf{y}_i}$ denote the k-nearest neighboring points of the overlapping points.

\emph{Pseudo Consistency Loss.}
We employ a cross-entropy based spatial consistency loss to sharpen the distribution of pseudo correspondences, it is defined as:
\begin{equation}
    \mathcal{L}_{pc}=-\frac1{|\mathbf{X_o}|}\sum_{\mathbf{x}_i\in\mathbf{X_o}}\sum_{j=1}^\mathbf{M}
    \mathbb{I}\left\{ j=\arg\max_{j'}\mathbf{M}_{i,j'}\right\}  \log\mathbf{M}_{i,j},
\end{equation}
where $\mathbb{I} \left\{\cdot \right\}$ is the indicator function and $\mathbf{M}$ is the matching matrix.

The overall loss is the sum of the three losses:
\begin{equation}
\mathcal{L} = \mathcal{L}_{gr} +  \mathcal{L}_{nc} +  \mathcal{L}_{pc}.
\end{equation}

The total loss is calculated in each iteration and time step, where each item has equal contribution to the final loss.

\begin{table*}
\caption{Results of different methods on 7Scenes, ICL-NUIM and KITTI, where KITTI results are not available for CEMNet.}
\setlength{\abovecaptionskip}{0pt}
\setlength{\belowcaptionskip}{0pt}
\begin{center}
        {   \resizebox{1\linewidth}{!}{
                \begin{tabular}{lcccc|cccc|cccccccc}
                    \toprule
                    \multicolumn{1}{c}{\multirow{2}{*}{Method}} &\multicolumn{4}{c|}{(a) 7Scenes} & \multicolumn{4}{c|}{(b) ICL-NUIM} & \multicolumn{4}{c}{(c) KITTI} \\
                    \cmidrule{2-3}   \cmidrule{4-5} \cmidrule{6-7}   \cmidrule{8-9} \cmidrule{10-11}   \cmidrule{12-13} \cmidrule{14-15}
                    &RMSE(\textbf{{R}})&MAE(\textbf{{R}})&RMSE(\textbf{{t}})&MAE(\textbf{{t}})&RMSE(\textbf{{R}})&MAE(\textbf{{R}})&RMSE(\textbf{{t}})&MAE(\textbf{{t}})
                    &RMSE(\textbf{{R}})&MAE(\textbf{{R}})&RMSE(\textbf{{t}})&MAE(\textbf{{t}})\\
                    \midrule
                    ICP $\left(\star\right)$ & 19.9166& 7.5760& 0.1127& 0.0310  & 10.1247& 2.1484& 0.3006&0.0693 & 19.7362& 4.0355& 1.8493&0.9124\\
                    FGR $\left(\star\right)$ &0.2724& 0.1380&0.0011&0.0006&3.0423& 1.9571&0.1275&0.0659& 8.2598&1.6986& 0.0794 & 0.0375\\ 
                    SDRSAC $\left(\star\right)$ &0.3501&0.2925&0.4997&0.4997&9.4074&7.8627&0.2477&0.2076&7.3050&1.6037&0.0688&0.0289\\
                    DCP $\left(\circ\right)$ & 7.5548&5.6991& 0.0411&0.0303&9.2142&6.5826&0.0191&0.0134 &10.3303 & 2.2953& 0.0985&0.0532 \\
                    IDAM$\left(\circ\right)$ &10.5306&5.6727 & 0.0539&0.0303&9.4539& 4.4153& 0.3040& 0.1385& 7.4124& 1.5751& 0.0620&0.0271\\
                    FINet$\left(\circ\right)$ &1.7824& 0.9038& 0.0094&0.0051&2.8731&1.1875&0.1273&0.0517 & 6.2106 & 1.4638 &0.0578 & 0.0348\\
                    FMR $\left(\triangleleft\right)$ &8.6999 & 3.6569& 0.0199 &0.0101&1.8282 &1.1085 & 0.0685 &0.0398&9.7362 &1.6809 &0.0848 &0.0305\\
                    CEMNet$\left(\triangleleft\right)$ & 0.1768& 0.0434&0.0012&0.0002& 0.8272&0.2316&\textbf{0.0021}&\textbf{0.0010}& -&- &- & -\\
                    RIENet $\left(\triangleleft\right)$ &\underline{0.0188} &\underline{0.0131}& \underline{0.0002} & \underline{0.0001} &\underline{0.1115}&\underline{0.0792}&0.0048&0.0034&\underline{5.5180}&\textbf{0.8840}&\underline{0.0457}&\textbf{0.0162}\\ 
                    \midrule 
                    Ours $\left(\triangleleft\right)$ &\textbf{0.0120}& \textbf{0.0079}&\textbf{0.0000}& \textbf{0.0000}& \textbf{0.0684}& \textbf{0.0517}& \underline{0.0033}& \underline{0.0026}& \textbf{2.0707}& \underline{1.0178}& \textbf{0.0302}& \underline{0.0180}\\
                    \bottomrule
            \end{tabular}}
        }
\end{center}
\label{table2}
\vskip -5pt
\end{table*}

\section{EXPERIMENTS}

\subsection{Experimental Settings}
We train our network end-to-end using PyTorch implementation with a 3090 GPU. During both training and testing, we utilize $3$ iterations and $3$ time steps. 
The Adam optimizer is employed with an initial learning rate of $10^{-3}$.
We compare our method with a range of other methods, including traditional approaches such as ICP~\cite{besl1992method}, FGR~\cite{zhou2016fast}, and SDRSAC~\cite{le2019sdrsac}, as well as learning-based supervised methods like DCP~\cite{wang2019deep}, IDAM~\cite{li2020iterative}, and FINet~\cite{xu2022finet}. Additionally, we evaluate our method against learning-based unsupervised methods such as FMR~\cite{huang2020feature}, CEMNet~\cite{jiang2021sampling}, and RIENet~\cite{shen2022reliable}.
Following~\cite{wang2019deep}, we measure anisotropic errors, including the root mean squared error (RMSE) and mean absolute error (MAE) of rotation and translation.

\subsection{Evaluation on Synthetic Dataset: ModelNet40}

\emph{ModelNet40.}
We conduct our evaluation on the ModelNet40 dataset~\cite{wu20153d}, which consists of 12,311 CAD models belonging to 40 different object categories. 
We randomly select 1,024 points from the outer surface of each model. During both training and testing, we apply rotations by sampling three Euler angle rotations within the range of ${\left[0^{\circ},45^{\circ}\right]}$. Additionally, translations are applied on each axis within the range of $\left[-0.5,0.5\right]$. We transform the source point cloud $\mathbf{X}$ using the sampled rigid transform and the task is to register it to the reference point cloud $\mathbf{Y}$. To perform partial-to-partial registration, we adopt the approach utilized in PRNet~\cite{wang2019prnet}. 
Subsequently, we retain the 768 points that are closest to this far point for each respective point cloud.

\emph{Comparison.}
Firstly, we train our model using the training set from ModelNet40, and evaluate its performance on the test set. It is important to note that both the training and test sets encompass point clouds from all 40 categories. As shown in Table~\ref{table1}(a), our method achieves the lowest error compared to both traditional and learning-based methods. Fig.~\ref{VIS}(a) showcases example results obtained using our approach.

We also evaluate the generalization capability of our approach to unseen categories. Specifically, we assess its performance on 20 new categories that have not been encountered during the model's training phase. Despite the new challenge posed by these unseen categories, our method consistently delivers excellent results. Table~\ref{table1}(b) summarizes the results, and the visual examples are presented in Fig.~\ref{VIS}(b).

Furthermore, we assess the performance of our model in the presence of noise, which is a common condition of real-world scenes. To simulate this scenario, we introduce random Gaussian noise with a standard deviation of 0.5, clipped to $\left[-1,1\right]$. 
As illustrated in Table~\ref{table1}(c), our method surpasses other methods in terms of performance. Additionally, Fig.~\ref{VIS}(c) provides the qualitative results.

\subsection{Evaluation on Indoor Dataset: 7Scenes, ICL-NUIM}
\emph{7Scenes.}
The 7Scenes~\cite{shotton2013scene} dataset is a widely used benchmark for registration in indoor environments.
Our model is trained on 6 categories (\emph{Chess}, \emph{Fires}, \emph{Heads}, \emph{Pumpkin}, \emph{Stairs} and \emph{Redkitchen}) and tested on the remaining category (\emph{Office}). The dataset is divided into 296 and 57 samples for training and testing.

\emph{ICL-NUIM.}
The ICL-NUIM dataset~\cite{choi2015robust} is a comprehensive collection of synchronized RGB-D video sequences acquired using a Kinect sensor. It encompasses a variety of indoor environments, such as offices, laboratories, and hallways. Before being divided into 1,278 samples for training and 200 samples for testing, the dataset undergoes augmentation.

\begin{figure*}[t]
\centering
\includegraphics[width=0.9\textwidth]{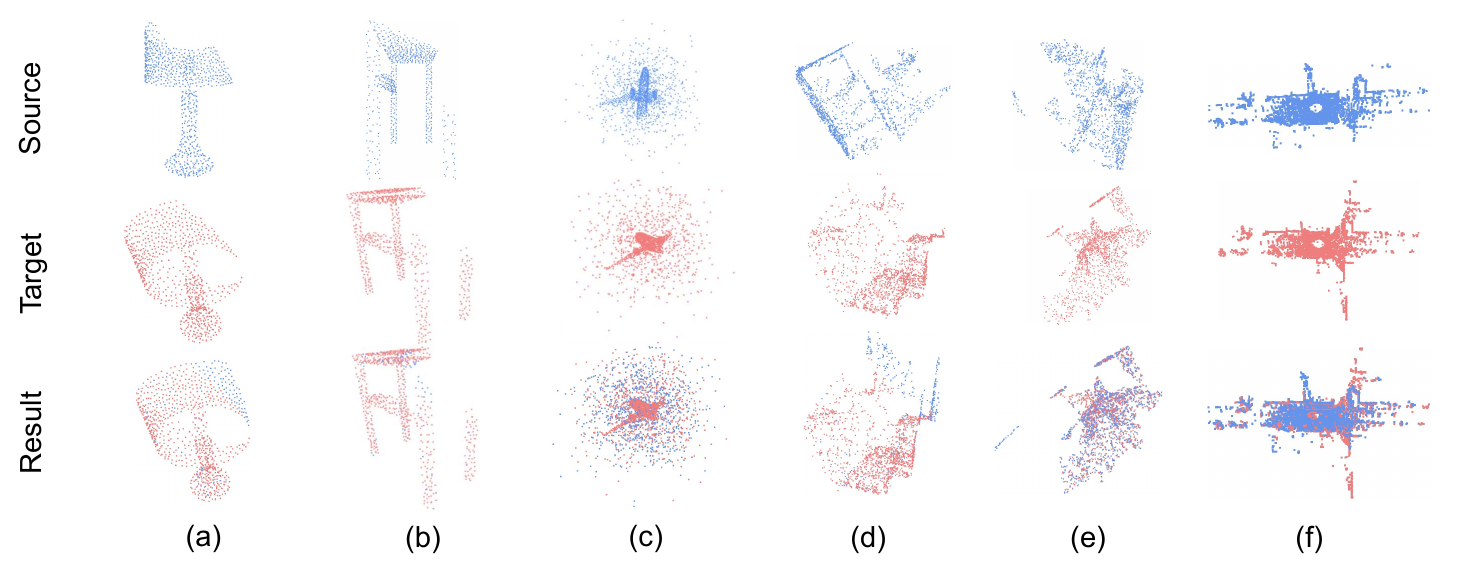} 
\caption{Qualitative registration results for (a, b, c) ModelNet40, (d) 7Scenes, (e) ICL-NUIM and (f) KITTI.}
\label{VIS}
\end{figure*}

\emph{Comparison.}
For two indoor datasets, we resample the source point clouds to 2,048 points and apply rigid transformation to generate the target point clouds. Then, we downsample the point clouds to 1,536 points to generate the partial data. 
Table~\ref{table2}(a) showcases the exceptional performance of our method on the 7Scenes dataset. Moreover, Table~\ref{table2}(b) reveals that our method achieves the best results in terms of the rotation metric for the ICL-NUIM dataset, while closely following CEMNet in the translation metric. 
To further illustrate the outcomes, Fig.~\ref{VIS}(d)(e) provides visual examples from the indoor datasets.

\subsection{Evaluation on Outdoor Dataset: KITTI}
\emph{KITTI.}
The typical outdoor scene dataset, KITTI~\cite{geiger2013vision} is used to evaluate our IFNet, which consists of LIDAR scans.
Following ~\cite{shen2022reliable}, the KITTI dataset comprises 11 sequences with ground truth pose. We use sequences 00-05 for training, 06-07 for validation, and 08-10 for testing. 
To construct pairwise point clouds, we combine the current frame with the 10th frame after it.

\emph{Comparison.}
We start by augmenting the dataset with random rotations, taking into account the initial pose. Following this, 
we voxelize the point clouds using a voxel size of 0.3m and randomly sample 2048 points from each voxelized representation.
The quantitative results are presented in Table~\ref{table2}(c). As observed, our method attains the top performance in terms of RMSE for both rotation and translation metrics. Additionally, our method ranks second in terms of MAE for both metrics, with RIENet taking the lead. For a visual representation of the outcomes, Fig.~\ref{VIS}(f) provides qualitative results.

\subsection{Ablation Studies}

\emph{Time Steps and Iteration times.}
In our ablation experiments, we train the models using the same categories on the ModelNet40 dataset and subsequently test them on unseen categories while also introducing 75\% missing points.
{
We conduct experiments with varying time steps and iteration times to assess the effectiveness of the feedback mechanism. The results of the experiment are displayed in Table~\ref{table3}.
For the purpose of optimizing efficiency and performance, we employ the settings of $s = 3$ and $t = 3$.
}


\emph{Feedback Transformer.}
As shown in Table~\ref{table4} (top), the results are mediocre when we merely concatenate high-level features with low-level features (Concat). However, when we utilize the $\textup{FT}$ module without positional embedding (FT \textit{w/o} pe), our results significantly improve, highlighting the positive impact of the $\textup{FT}$ module. Furthermore, employing the 3D coordinates as positional embedding (FT \textit{w/} xyz) leads to further improvements. Moreover, using our proposed geometry-aware descriptor ($\textup{gad}$) as positional embedding (FT \textit{w/} gad) results in even better performance.

\begin{figure}[t]
\begin{center}
\includegraphics[width=0.9\linewidth]{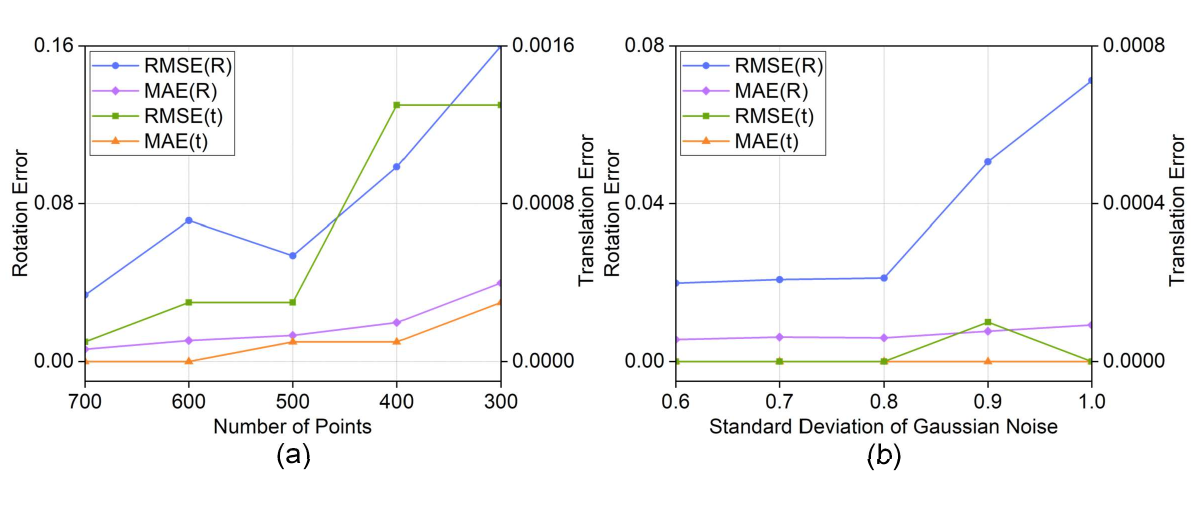}
\end{center}
\vskip -10pt
\caption{Robustness test. (a) Errors under different number of points. (b) Errors under different noise levels.}
\label{robust}
\vskip -10pt
\end{figure}

\begin{table}[t]

    \caption{Ablation studies of different time steps and iteration times.}
    \resizebox{\linewidth}{!}{%
    \begin{tabular}{lcccccccccccccccc}
        \toprule
         Method      & RMSE(\textbf{{R}}) & MAE(\textbf{{R}}) & RMSE(\textbf{{t}}) & MAE(\textbf{{t}}) & time(s) \\
        \midrule
        $s=1$       & 0.2170          & 0.0230         & 0.0012          & 0.0003      & \textbf{33.62}        \\
        $s=2$           & 0.1376       & 0.0180   & 0.0008  & 0.0003   & 36.38       \\
        $s=3$      & 0.0340         & 0.0061         & \textbf{0.0002}       & \textbf{0.0000}  & 45.96\\
        $s=4$          & \textbf{0.0337}          & \textbf{0.0059}         & \textbf{0.0002}      & \textbf{0.0000}      & 57.16  \\
        \midrule
        $t=1$       & 0.1559          & 0.0135         & 0.0010          & 0.0002      & \textbf{32.94}        \\
        $t=2$           & 0.1044       & 0.0095   & 0.0008  & 0.0002   & 39.01       \\
        $t=3$      & \textbf{0.0340}         & \textbf{0.0061}         & \textbf{0.0002}       & \textbf{0.0000}  & 45.96\\
        $t=4$          & 0.0353          & 0.0066         & \textbf{0.0002}      & \textbf{0.0000}      & 51.84  \\

         \bottomrule
    \end{tabular}%
    }
    \vspace{-0.37cm}
	\label{table3}
\end{table}

\begin{table}[t]

    \caption{Ablation studies of the Feedback Transformer and loss functions.}
    \resizebox{\linewidth}{!}{%
    \begin{tabular}{lcccccccccccccccc}
        \toprule
         \textup{Strategy}      & RMSE(\textbf{{R}}) & MAE(\textbf{{R}}) & RMSE(\textbf{{t}}) & MAE(\textbf{{t}}) \\
        \midrule
        Concat       & 1.7259          & 0.0559         & 0.0114          & 0.0007              \\
        FT \textit{w/o} pe           & 0.2864       & 0.0208   & 0.0030  & 0.0002          \\
        FT \textit{w/} xyz      & 0.0590         & 0.0074         & 0.0004       & \textbf{0.0000} \\
        FT \textit{w/} gad         & \textbf{0.0340}          & \textbf{0.0061}         & \textbf{0.0002}      & \textbf{0.0000}        \\
        \midrule
        $\mathcal{L}_{gr}$ & 0.9617          & 0.0386         & 0.0088          & 0.0005              \\
        $\mathcal{L}_{gr}+\mathcal{L}_{pc}$           & 0.1167       & 0.0193   & 0.0007  & 0.0003        \\
        $\mathcal{L}_{gr}+\mathcal{L}_{pc}+\mathcal{L}_{nc}$      & \textbf{0.0340}          & \textbf{0.0061}         & \textbf{0.0002}      & \textbf{0.0000}   \\
        
         \bottomrule
    \end{tabular}%
    }
    \vspace{-0.37cm}
	\label{table4}
\end{table}

{
\emph{Loss Functions.}
In our experiments, we train our model with the combination of the global registration loss ($\mathcal{L}_{gr}$), the neighborhood consistency loss ($\mathcal{L}_{nc}$) and the pseudo consistency loss ($\mathcal{L}_{pc}$). 
From Table~\ref{table4} (down), it can be seen that each loss has a positive effect on performance.
}

{
\emph{Time Efficiency.}
We calculate the time efficiency of the different methods. 
The inference time of our method is $146$ms, while the cost of the other methods are ICP ($8$ms), FINet ($26$ms), DCP ($30$ms), IDAM ($49$ms), FGR ($55$ms), RIENet ($62$ms),  CEMNet ($295$ms), FMR ($361$ms), and SDRSAC ($22,416$ms), respectively.
}

\emph{Robustness Analysis.}
To showcase the robustness of our model to the number of points, we test it on varying numbers of points within the range of [300, 700].
The results are displayed in Figure~\ref{robust}(a), where it can be observed that our IFNet maintains robust performance even as the number of points decreases.
Additionally, we also test at varying noise levels within the range of [0.6, 1.0]. As shown in Figure~\ref{robust}(b), our IFNet consistently achieves comparable performance across these different noise levels.


{
\emph{Lower Overlap.}
To evaluate the performance in a low overlap ratio, we independently place the far point for the two point clouds. The remaining pre-processing steps remain consistent with ~\cite{wang2019prnet}. 
Table~\ref{table5} displays the results, indicating that our method outperforms the baseline method RIENet~\cite{shen2022reliable} in both same and unseen conditions.
}

\begin{table}[t]

    \caption{Comparison of our method with RIENet under lower overlap.}
    \resizebox{\linewidth}{!}{%
    \begin{tabular}{lcccccccccccccccc}
        \toprule
         Condition   & Method   & RMSE(\textbf{{R}}) & MAE(\textbf{{R}}) & RMSE(\textbf{{t}}) & MAE(\textbf{{t}})  \\
        \midrule
        Same       & RIENet          & 0.2204       & 0.0422          & 0.0007      & 0.0003        \\
        Same           & Ours       & \textbf{0.1840}  & \textbf{0.0282}  & \textbf{0.0005}   & \textbf{0.0001}       \\
        \midrule
        Unseen      & RIENet         & 0.3190         & 0.0553       & 0.0008  & 0.0003\\
        Unseen          & Ours          & \textbf{0.0885}         & \textbf{0.0149}      & \textbf{0.0003}     & \textbf{0.0000}  \\ 
        
         \bottomrule
    \end{tabular}%
    }
    \vspace{-0.37cm}
	\label{table5}
\end{table}

\section{CONCLUSIONS}

We propose IFNet, an end-to-end unsupervised method that employs an iterative feedback mechanism for 3D point cloud registration. 
Our IFNet consists of multiple Feedback Registration Block (FRB) modules.
By incorporating the Feedback Transformer in the FRB module, we can extract more representative and informative low-level features by leveraging high-level information. 
Moreover, we propose the geometry-aware descriptor, which serves as a positional embedding for the Feedback Transformer, enabling our model to fully exploit geometric information. 
By the stacking of FRB modules, the outputs are gradually refined across time steps, ultimately leading to the precise estimation of rigid transformations.
Extensive experiments on 
the ModelNet40, 7Scenes, ICL-NUIM, and KITTI benchmarks demonstrate that IFNet achieves superior performance.

\addtolength{\textheight}{-2cm}   

\bibliographystyle{IEEEtran}
\balance

\end{document}